# CLASSES FOR FAST MAXIMUM ENTROPY TRAINING

*Joshua Goodman*

Microsoft Research
Redmond, Washington 98052, USA
joshuago@microsoft.com
http://research.microsoft.com/~joshuago

## ABSTRACT

Maximum entropy models are considered by many to be one of the most promising avenues of language modeling research. Unfortunately, long training times make maximum entropy research difficult. We present a novel speedup technique: we change the form of the model to use classes. Our speedup works by creating two maximum entropy models, the first of which predicts the class of each word, and the second of which predicts the word itself. This factoring of the model leads to fewer non-zero indicator functions, and faster normalization, achieving speedups of up to a factor of 35 over one of the best previous techniques. It also results in typically slightly lower perplexities. The same trick can be used to speed training of other machine learning techniques, e.g. neural networks, applied to any problem with a large number of outputs, such as language modeling.

## 1. INTRODUCTION

Maximum entropy models [1] are perhaps one of the most promising techniques for language model research. These techniques allow diverse sources of information to be combined. For each source of information, a set of constraints on the model can be determined, and then, using an algorithm such as Generalized Iterative Scaling (GIS), a model can be found that satisfies all of the constraints, while being as smooth as possible. However, training maximum entropy models can be extremely time consuming, taking weeks, months, or more. We show that by using word classing, the training time can be significantly reduced, by up to a factor of 35. In particular, we change the form of the model. Instead of predicting words directly, we first predict the class that the next word belongs to, and then predict the word itself, conditioned on its class. The technique used is actually more general: it can be applied to any problem where there are a large number of outputs to predict, with language modeling being just one example. Furthermore, the technique applies not only to maximum entropy models, but to almost any machine learning technique for predicting probabilities that is slowed by a large number of outputs, including many uses of decision trees and neural networks.

In this paper, we first give a very brief introduction to maximum entropy techniques for language modeling. We then go on to describe our speedup. After this, we describe previous research in speeding up maximum entropy training, and compare it to our technique. Next, we give experimental results, showing both the increased speed of training, and a slight reduction in perplexity of the resulting models. Finally, we conclude with a short discussion of how these results can be applied to other machine learning techniques and to other problems.

We begin with a very quick introduction to language models in general, and maximum entropy-based language models in particular. Language models assign probabilities to word sequences $P(w_1...w_n)$. Typically, this is done using the trigram approximation:

$$P(w_1...w_n) = \prod_{i=1}^{n} P(w_i | w_1...w_{i-1}) \approx \prod_{i=1}^{n} P(w_i | w_{i-2} w_{i-1})$$

Maximum entropy models for language modeling [2] do not necessarily use the n-gram approximation and can in principle condition on arbitrary length contexts. The general form of a conditional maximum entropy model is as follows:

$$P'(w | w_1...w_{i-1}) = \frac{\exp\left(\sum_j \lambda_j f_j(w, w_1...w_{i-1})\right)}{Z_\lambda(w_1...w_{i-1})}$$

The $\lambda_j$ are real-valued constants learned in such a way as to optimize the perplexity of training data. $Z_\lambda(w_1...w_{i-1})$ is a normalizing constant so that the sum of all probabilities is 1, simply set equal to $\sum_w \exp\left(\sum_j \lambda_j f_j(w, w_1...w_{i-1})\right)$. The $f_j$ represent a large set of indicator functions that always have the value 1 or 0. For instance, we could use $f_j(\text{"Tuesday"}, w_1...w_{i-1}) = 1$ if $w_{i-1} = \text{"on"}$ and $w_{i-2} = \text{"meet"}$ (and, implicitly, otherwise 0). If $\lambda_j$ were given a positive value, then the probability of "Tuesday" in the context of "meet on" would be raised. By making many indicator functions of this type, we could capture all of the information captured by a trigram. Similarly, we could make a bigram indicator function with $f_j(\text{"Tuesday"}, w_1...w_{i-1}) = 1$ if $w_{i-1} = \text{"on"}$. Or we could make a unigram indicator function with the simple $f_j(\text{"Tuesday"}, w_1...w_{i-1})=1$ for all $w_1...w_{i-1}$. In principle, any set of indicator functions that depends on $w, w_1...w_{i-1}$ can be used, including n-grams, caching, skipping n-grams, word triggers, etc.

The optimal $\lambda$-values must be learned. The algorithm – Generalized Iterative Scaling [1] – for optimizing their values (based on some set of training data) can be very slow. It requires many iterations, and at each iteration, it involves a loop over all words in the training data. We give here a very rough sketch of the algorithm, with only the inner loop presented in any detail.

The inner loop of this code, and the most time-consuming part is lines 4 to 12. Notice that the inner loop contains several loops over all words in the vocabulary (lines 4, 7 and 8). Notice

```
1   For each iteration
2     observed[1..# of indicators] ← {0}
3     For i = 1 to |training data|
4       For each word w in Vocabulary
5         unnormalized[w] ← exp($\sum_j \lambda_j f_j(w, w_1...w_{i-1})$)
6       Next w
7       z ← $\sum_w$ unnormalized[w]
8       For each word w in Vocabulary
9         For each j such that $f_j(w, w_1...w_{i-1}) \neq 0$
10          observed[j] += $f_j$ × unnormalized[w]/z
11        Next j
12      Next w
13    Next i
14    For each indicator $f_j$
15      re-estimate $\lambda_j$ using observed[j]
16    Next j
17  Next iteration
```

that the sum in line 5 is typically bounded by the number of different types of indicator functions. In particular, a given system will typically have only a few types of indicators – e.g. unigram, bigram, and trigram – and typically, for each of these types, and for a given word w, and history, there can be only one non-zero indicator function. This means that the sum in line 5 and the loop in line 9 are bounded by the number of types of indicator function. Overall, then, the inner loop of lines 4 to 12 is typically bounded by the number of types of indicator function, times the vocabulary size. This means that decreasing the vocabulary size leads to a decrease in the runtime of the inner loop. Certain types of indicator functions (e.g. triggers) and optimizations to line 5 (summing only over non-zero $f_j$) change the exact analysis of run-time, but not the intuition that inner-loop run-time is roughly proportional to vocabulary size.

## 2. CLASS-BASED SPEEDUP

We now describe our speedup. We assign each word in the vocabulary to a unique class. For instance, cat and dog might be in the class of ANIMAL, while Tuesday and Wednesday might be in the class of WEEKDAY. Next, we observe that

$$P(w | w_1...w_{i-1}) = P(class(w) | w_1...w_{i-1}) \times P(w | w_1...w_{i-1}, class(w))$$

This equality holds because each word is in a single class, and is easily proven. Conceptually, it says that we can decompose the prediction of a word given its history into prediction of its class given the history, and the probability of the word given the history and the class. For "true" probabilities, this equality is exact. If the probabilities are not true, but instead are, for instance, the results of estimating a model, or are smoothed, or are the results of computing a maximum entropy model, the equality will not be exact, but will be a very good approximation. Indeed, the approximation is so good that typically classing is used to lower the perplexity of models.

This decomposition is the basis for our technique. Rather than create a single maximum entropy model, we create two different models, the first of which predicts the class of a word given its context $P(class(w)/w_1...w_{i-1})$, and the second of which predicts a word given its class and its context, $P(w/w_1...w_{i-1}, class(w_i))$. The process of training each of these two models is completely separate. If we have 100 classes in our system, then the inner loop of the training code for predicting the class is bounded by a factor of 100, rather than a factor of the vocabulary size. Thus, this model can be computed relatively quickly.

Next, consider the unigram, bigram and trigram indicators for a class-based model. An example unigram indicator would be $f_j$("Tuesday", $w_1...w_{i-1}$, $class(w_i)$)=1 if $class(w_i)$=WEEKDAY. For the bigram indicator, we would have $f_j$("Tuesday", $w_1...w_{i-1}$, $class(w_i)$) = 1 if $class(w_i)$=WEEKDAY and $w_{i-1}$ = "on"; and a trigram indicator would be $f_j$("Tuesday", $w_1...w_{i-1}$, $class(w_i)$) = 1 if $class(w_i)$=WEEKDAY and $w_{i-1}$="on" and $w_{i-2}$="meet".

Notice an important fact: for a word w not in the same class as $w_i$, $P(w/w_1...w_{i-1}, class(w_i))$=0. This means that we can modify the loops of lines 4, 7 and 8 to loop only over those words w such that w is in the same class as $w_i$. Now, if each class has 100 words, then the run time of the inner loop is bounded by a factor of 100. (If we were to explicitly perform the computations, the unigram $\lambda$'s for words w not in $class(w_i)$ would be set to -∞, and the unnormalized probabilities would be 0, leading to no contribution in lines 7 and 10).

Consider a hypothetical example, with a 10,000 word vocabulary, 100 classes and 100 words per class. The inner loop of the standard training algorithm would require time proportional to 10,000. Alternatively, we can use the class-based speedup. Both the inner loop for learning the class model, and the inner loop for running the word-given-class model are bounded by a factor of 100, leading to an overall hypothetical improvement of 10,000/(100+100)=50.

We can extend this result to 3 or more levels, by predicting first a super-class, e.g. NOUN, and then a class, e.g. WEEKDAY, and finally the word, Tuesday. Such a decomposition will further reduce the maximum number of indicator functions, but, since there is some overhead to each level, we have not found improvements by extending beyond 3 levels.

## 3. PREVIOUS RESEARCH

Maximum entropy has been well studied. [1] gives the classic Generalized Iterative Scaling algorithm, although in a form suitable for joint probabilities, as opposed to the conditional probabilities given here, and is somewhat dense; [2] is a classic introduction to the use of maximum entropy models for language modeling, but despite the fact that [2] uses conditional probabilities, most of the discussion is of joint probabilities.

[2] has previously used a simple form of classes with maximum entropy-based language models. However, they were used only as *conditioning* variables; i.e., indicator functions like $f_j(w/w_{i-1})$ = 1 if w=y and $class(w_{i-1})$=x were used. They were not used for *predicting* outputs, and thus did not lead to speedups.

Word classes have, of course, been used extensively in language modeling, including [2][3][4][5]. However, previous research has focused mostly on improving perplexity or reducing language model size, and never to our knowledge for increasing speed. Note that we have previously used a model form very similar to the one used here for reducing language model size, by up to a factor of 3, at the same perplexity [5].

There have been three noteworthy previous attempts to speed up maximum entropy models: unigram caching, Improved Iterative Scaling (IIS) [6], and cluster expansion [7][8].

Unigram caching makes use of the following observation: most bigram and trigram indicators are not used in practice (e.g. if the string "York Francisco" never occurred in the training data, then there will not be any bigram indicators for that case). On the other hand, all possible unigram indicators typically *are* used. This means that typically, the vast majority of indicator functions that are non-zero for a given context are unigram indicators; also notice that these unigram indicators are independent of context, meaning computation can be easily shared. In unigram caching, the effect of the unigram indicators is pre-computed and the computations of the inner loop are rearranged so that they depend only on those non-unigram indicators that take a non-zero value. In practice, the number of non-zero indicators still tends to be proportional to the vocabulary size (since the number of non-zero bigrams, trigrams, and similar indicator-functions for a given history is bounded by the vocabulary size).

We have implemented unigram caching and it leads to considerable speedups over the naïve implementation. Our 35 times speedup is a speedup over unigram caching. Our technique can be used with or without unigram caching, but because of some extra overhead involved in unigram caching, and because our technique drastically reduces the number of unigrams, it is usually best to use our technique without unigram caching.

In Improved Iterative Scaling [6], a different update technique is used. It introduces additional overhead that slows down the time for each iteration of the iterative scaling algorithm, but allows larger steps to be taken at each time, leading to fewer iterations, and overall faster performance. It also introduces additional memory overhead and coding complexity. The main benefits from improved iterative scaling come from certain models in which the total number of indicator functions that can be true for a certain time is highly variable. The learning speed of Generalized Iterative Scaling is inversely proportional to the value of $\max_w \max_i \sum_j f_j(w, w_1...w_{i-1})$.

GIS uses the maximum of this value to slow learning, while IIS slows learning on a case-by-case basis. In some models, this sum can be very different for different $w, i$. In particular, models using caching and triggering techniques can lead to these different numbers of active indicators. In other models, such as n-gram-style models, there is a fixed maximum for the number of non-zero indicators. In these models, IIS would lead to little or no reduction in the number of iterations of iterative scaling, and, because of the additional overhead for each iteration, might actually lead to a slowdown.

The last technique we consider is the most powerful one, cluster expansion, introduced in [7] and expanded in [8]. Cluster expansion can be regarded as a natural extension of unigram caching to n-grams. Consider a simple trigram model. With some straightforward rearrangement of the equations, for two trigrams with a common bigram, most of the computation can be shared. In [8], this technique is extended to handle cases in which there is limited interaction between hierarchical constraints, and still achieves good speedups (a factor of 15.) However, as [7] concludes, cluster expansion "is limited in its usefulness… When the number of interacting constraints is large…the cluster expansion is of little use in computing the exact maximum entropy solution." We believe that the same conclusion applies to [8]. In particular, a simple model combining bigram 1-back, 2-back, …, 5-back constraints would probably show no, or only small, gains from the techniques of [7] or [8], while for our technique, the gains would be about the same.

In theory, our speedup can be used in conjunction with IIS, unigram caching, or cluster expansion. However, in conjunction with unigram caching, in experiments, it typically leads to only small speed improvements, or sometimes actual slowdowns (because unigram caching introduces overhead in other parts of the algorithm). Similarly, we suspect that with cluster expansion speedup might be limited. We have not tested our algorithm with IIS, but in principle, there is no reason they could not be combined, and we guess the combination would work well.

## 4. RESULTS

We ran our experiments using four different learning techniques: simple GIS, GIS with unigram caching, GIS with a two level clustering, and GIS with a three level clustering. We ran on four different sizes of training data. The model used is a "skipping, classing" model with the following types of indicator functions, where *W*, *Y*, and *Z* are variables filled in for specific instances of the indicator functions.

$f_j^{\text{unigram}}(w, w_1...w_{i-1})=1$ if $w=W$

$f_j^{\text{class-bigram}}(w, w_1...w_{i-1})=1$ if $w=W$ and $class(w_{i-1})=Z$

$f_j^{\text{class-skip-bigram}}(w, w_1...w_{i-1})=1$ if $w=W$ and $class(w_{i-2})=Z$

$f_j^{\text{bigram}}(w, w_1...w_{i-1})=1$ if $w=W$ and $w_{i-1}=Z$

$f_j^{\text{skip-bigram}}(w, w_1...w_{i-1})=1$ if $w=W$ and $w_{i-2}=Z$

$f_j^{\text{class-trigram}}(w, w_1...w_{i-1})=1$ if $w=W$ and $class(w_{i-1})=Z$ and $class(w_{i-2})=Y$

$f_j^{\text{class-bigram-skip-bigram}}(w, w_1...w_{i-1})=1$ if $w=W$ and $class(w_{i-1})=Z$ and $w_{i-2}=Y$

$f_j^{\text{bigram-class-skip-bigram}}(w, w_1...w_{i-1})=1$ if $w=W$ and $w_{i-1}=Z$ and $class(w_{i-2})=Y$

We used all and only indicator functions where there were at least three matching cases in the training data. We found our word classes by using a top-down splitting algorithm that attempted to minimize entropy loss, as described in [9]. We used different numbers of classes for different purposes. For the two-level splitting, we used approximately 60-250 classes. For the three-level splitting, we used approximately 8-30 classes for the first level, and 100-2000 classes for the second level. In all cases, we optimized the number of classes by running one iteration of training with varying numbers of classes, and picking the fastest. The classes used in the indicator functions are not typically the same as the classes used in our factoring; for the indicator classes, we used 64 classes. We linearly interpolated each maximum entropy model with a trigram model, to smooth and avoid zero probabilities. Our technique interpolated with a trigram model reduced overall perplexity from 1% to 5% versus a maximum entropy model without our technique interpolated with the same trigram model; we were not able to run the baseline perplexity at 10,000,000 words, because the version without our speedups was too slow. We used subsets of Wall Street Journal data, building the classes from scratch at each size, and using the 60,000 most common words in the training data, or all words, if there were fewer than 60,000 unique words in the training data.

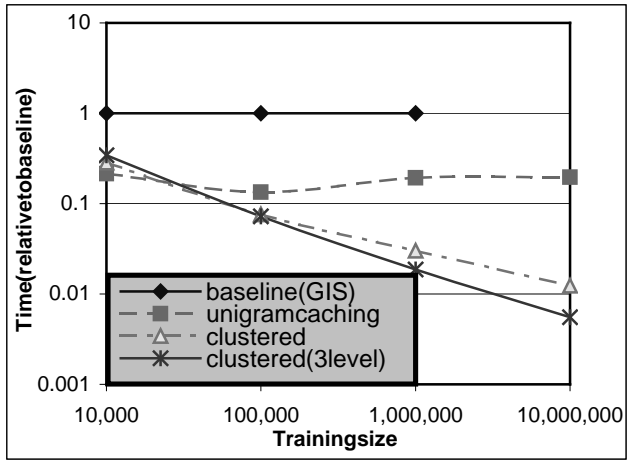

**Figure 1: Speedup Results**

Figure 1 shows our results, giving relative speeds. Notice that we achieve a speedup of up to a factor of 35 over the unigram caching result. We believe this is the largest speedup reported. Notice also that at the smallest data size, the classing methods actually result in minor slowdowns compared to unigram caching, but that as the training data size increases, the speedup from our technique also rapidly increases.

## 5. DISCUSSION

We have discussed our speedup technique in the context of training. However, it can in many cases also be used for testing. In particular, if in the test situation one needs the probabilities of most or all words in a particular context, our speedup will not be helpful. On the other hand, if one were to use maximum entropy models to rescore n-best lists, the speedup would work just as well for testing as for training. In the case of, say, rescoring lattices, the speedup would be helpful, as long as the lattices did not allow for too many words in each context.

Notice that our speedup technique could be applied to a variety of other problems and to a variety of other learning methods. In particular, there is nothing in particular specific to language modeling in our speedup technique, except that we are predicting the probabilities of a large number of outputs (possible next words). Any other problem predicting the probabilities of a large number of outputs could benefit from these methods. Similarly, there is little that is specific to maximum entropy models in our technique. For instance, consider training a neural network to learn the probabilities of 10,000 outputs. Each step of training would require back-propagating 9,999 zeros, and one 1. Alternatively, one could place the outputs into 100 classes. A first network could be trained to learn the class-probabilities. Each step of training would require back-propagating 99 zeros and one 1. Next, we would learn 100 neural networks, for predicting the probability of the outputs given the class, one neural network for each class, predicting a probability for each output in that class. Network $i$ would learn the conditional probabilities of outputs in class $i$ given that class $i$ is correct. Each step of training would need only train the network corresponding to the correct class, meaning that again, only 99 zeros and one 1 would need to be back-propagated. Presumably the number of hidden units of these smaller networks predicting only 100 values (for the class, or the outputs given the class) would also be much smaller than the number of hidden units of a network for predicting 10,000 outputs directly.

Similarly, there are at least two ways to train decision trees to handle large numbers of outputs (train decision trees with many outputs at each leaf, or train a binary decision tree for each possible output and normalize). Again, in both of these cases, our method can be applied.

More generally, almost any learning algorithm that is slowed at training time when there are a large number of outputs could benefit from our approach. Similarly, any algorithm slowed at test time by a large number of outputs, but used in a situation in which only a few of those outputs are needed, would benefit.

Our technique is an extremely promising one. Although it is only an approximation, rather than an exact technique, it is one that both theoretically and empirically reduces perplexity; it adds very little complexity to coding; it leads to perhaps the largest reported speedups – a factor of 35; these speedups are largest when needed most, on large, complex problems; it can be applied independently of the form of the model; and it can be applied both to other learning algorithms and to other problem domains. We are hopeful that others will use it, both for maximum entropy modeling applied to language modeling and in many other fields.